\documentclass[preprints,article,accept,pdftex,moreauthors]{Definitions/mdpi} 
\firstpage{1} 
\pubvolume{1}
\issuenum{1}
\articlenumber{0}
\pubyear{2022}
\copyrightyear{2022}
\datereceived{} 
\dateaccepted{} 
\datepublished{} 
\hreflink{https://doi.org/} 

\Title{Human Evaluation of English--Irish Transformer-Based NMT} 

\TitleCitation{Human Evaluation of English--Irish Transformer-Based NMT} 

\Author{S\'{e}amus Lankford $^{1,2,}$*$^{,\dagger}$, Haithem Afli $^{2}$ and Andy Way $^{1}$}

\AuthorNames{Séamus Lankford, Haithem Afli and Andy Way}

\AuthorCitation{Lankford, S.; Afli, H.; Way,~A.}

\address{%
$^{1}$ \quad ADAPT Centre, Dublin City University, D09 Y074 Dublin, Ireland; andy.way@adaptcentre.ie\\
$^{2}$ \quad Munster Technological University, T12 P928 Cork, Ireland; haithem.afli@mtu.ie}

\corres{Correspondence: seamus.lankford@adaptcentre.ie}

\firstnote{Current address: ADAPT Centre, School of Computing, Dublin City University, D09 Y074 Dublin, Ireland.}

\abstract{
In this study, a human evaluation is carried out on how hyperparameter settings impact the quality of Transformer-based Neural Machine Translation (NMT) for the low-resourced English--Irish pair. SentencePiece models using both Byte Pair Encoding (BPE) and unigram approaches were appraised. Variations in model architectures included modifying the number of layers, evaluating the optimal number of heads for attention and testing various regularisation techniques.  The greatest performance improvement was recorded for a Transformer-optimized model with a 16k BPE subword model. Compared with a baseline Recurrent Neural Network (RNN) model, a Transformer-optimized model demonstrated a BLEU score improvement of 7.8 points. When benchmarked against Google Translate, our translation engines demonstrated significant improvements. Furthermore, a quantitative fine-grained manual evaluation was conducted which compared the performance of machine translation systems. Using the Multidimensional Quality Metrics (MQM) error taxonomy, a human evaluation of the error types generated by an RNN-based system and a Transformer-based system was explored. Our findings show the best-performing Transformer system significantly reduces both accuracy and fluency errors when compared with an RNN-based model.}

\begin{document}





\section{Introduction}
A new era of high-quality translations has been heralded with the advent of NMT. Given that large datasets are a prerequisite for high-quality NMT, these improvements are not always evident in the translation of low-resource languages. In the context of such languages, which suffer from a sparsity of data, alternative approaches must be adopted. 

Developing applications and models to address the challenges of low-resource language technology is an important part of this research. This technology incorporates new methods, which reduce the impact that data scarcity has on the digital engagement of low-resource languages. One approach is to use a mechanism that helps NMT systems to learn from unlabeled data using dual-learning~\citep{he2016dual, ahmadnia2019augmenting}.
  
Out-of-the-box NMT systems, trained on English--Irish data, have been shown to achieve a lower translation quality compared with using a tailored SMT system~\citep{dowling2018smt}. It is in this context that further research is required in the development of NMT for low-resource languages, and the Irish language in particular.

\textls[-15]{Most research on the choice of subword models has focused on high-resource languages~\citep{ding2019call, gowda2020finding}}. Translation, by its nature, requires an open vocabulary and the use of subword models aims to address the fixed-vocabulary problem associated with NMT. Rare and unknown words are encoded as sequences of subword units. By adapting the original BPE algorithm~\citep{gage1994new}, the use of BPE submodels can improve translation performance~\citep{sennrich2015neural,kudo2018subword}. In the context of developing models for English-to-Irish translation, there were no clear recommendations on the choice of subword model types. Character-based models were briefly explored due to their simplicity and reduced memory requirements. However, they were not considered suitable given that most single characters do not carry meaning in the English and Irish languages. Therefore, one of the objectives of our research is to identify which type of subword model performs best in this low-resource scenario.

An important goal of this study is to extend our previous work~\citep{lankford2021transformers} by providing a human evaluation (HE) and comparison of EN→GA machine translation (MT) on systems that use either a baseline RNN architecture or a subword-model optimized Transformer~model. 

This paper describes the context in which our research was conducted and provides a background of the types of available architecture in Section \ref{s2}. A detailed overview of our approach is outlined in Section \ref{s3}, where we provide details of the data and parameters used in our NMT systems. The empirical results, using both automatic metrics and a human evaluation, are presented in Section \ref{s4}. Finally, our findings are discussed and possibilities for future work are described in Section \ref{s6}.

\section{Background\label{s2}}
   
Native speakers of low-resource languages are often excluded from useful content since, more often than not, online content is not available to them in their language of choice. This digital divide experienced by second-language speakers has been well-documented in the research literature ~\citep{macfarlane2008responses, alam2015digital}.

Research on MT in low-resource scenarios seeks to directly addresses this challenge of exclusion via pivot languages~\citep{liu2018pivot}, and indirectly, via domain adaptation of models ~\citep{ghifary2016deep}. Consequently, research efforts focusing on NMT~\citep{bahdanau2014neural, cho2014properties} have resulted in a state-of-the-art (SOA) performance being attained for multiple language pairs~\citep{bojar-etal-2017-findings, bojar-etal-2018-findings}. The Irish language is a primary example of a low-resource language that will benefit from this research. NMT involving Transformer model development will improve performance in specific domains of low-resource languages. 

\subsection{Hyperparameter Optimization}

Hyperparameters are employed  to customize machine learning models such as translation models. It has been shown that machine learning performance may be improved through hyperparameter optimization (HPO) rather than just using default settings~\citep{sanders2017informing}. The principal methods of HPO are Grid Search~\citep{montgomery2017design} and Random Search~\citep{bergstra2012random}. 

\subsubsection{RNN}

The tasks of natural language processing (NLP), speech recognition and MT are often performed by RNNs. This architecture enables previous outputs to be used as inputs while having hidden states. In the context of MT, such neural networks were ideal due to their ability to process inputs of any length. Furthermore, the model sizes do not necessarily increase with the input size. Commonly used variants of RNN include Bidirectional (BRNN) and Deep (DRNN) architectures. However, the problem of vanishing gradients coupled with the development of attention-based algorithms often leads to Transformer models performing better than RNNs.

\subsubsection{Transformer}
The greatest improvements have been demonstrated when either the RNN or the CNN architecture is abandoned completely and replaced with an attention mechanism creating a much simpler and faster architecture known as Transformer. Experiments in MT tasks show such models are better in quality due to greater parallelization while requiring significantly less time to train~\citep{vaswani2017attention}. 

Transformer models use attention to focus on previously generated tokens. The approach allows for models to develop a long memory, which is particularly useful in the  domain of language translation. Performance improvements to both RNN and CNN approaches may be achieved through the introduction of such attention layers in the translation architecture.

\subsection{SentencePiece}

Designed for NMT, SentencePiece is a language-independent subword tokenizer that provides an open-source C++ and a Python implementation for subword units. An attractive feature of the tokenizer is that SentencePiece directly trains subword models from raw sentences~\citep{kudo2018SentencePiece}.

\subsection{Human Evaluation}

Human evaluation, within NLP and MT, is a topic of growing importance, which often has its own dedicated research track or workshop at major conferences \citep{belz2021proceedings}. This focus has resulted in many publications in the area of HE that relate to MT \citep{toral2018attaining, castilho2017neural} and it has particularly benefited the evaluation of low-resource languages \citep{bayon2019evaluating, imankulova2019exploiting}.

The best practice for the HE of MT has been published in the form of a series of recommendations \citep{laubli2020set}. As part of our research, we adopted these recommendations, which are in line with similar EN-GA HE studies at the ADAPT centre \citep{dowling2020human}. Specifically, these recommendations encourage the use of professional translators, evaluation at the document level and assessments of both fluency and accuracy. Original source texts were also used in the training and test data.

These recommendations have been complemented by a fine-grained human analysis, which uses both a Scalar Quality Metric (SQM) and the MQM.

\section{Proposed Approach\label{s3}}

Considerable performance improvements have been achieved using the HPO of RNN models in low-resource settings. One of the key research questions, evaluated as part of this study, is to identify the extent to which such optimization techniques may be applied to low-resource Transformer models. Evaluations included modifying the number of attention heads, changing the number of layers and experimenting with regularization techniques such as dropout and label smoothing. Most importantly, the choice of subword model type and the vocabulary size are evaluated. Furthermore, previous research  focuses on using an automatic evaluation of performance, whereas we propose combining a HE approach with automatic metrics.

\begin{figure}[H]
\begin{adjustwidth}{-\extralength}{0cm}
\centering 
 \includegraphics[width=17.8cm]{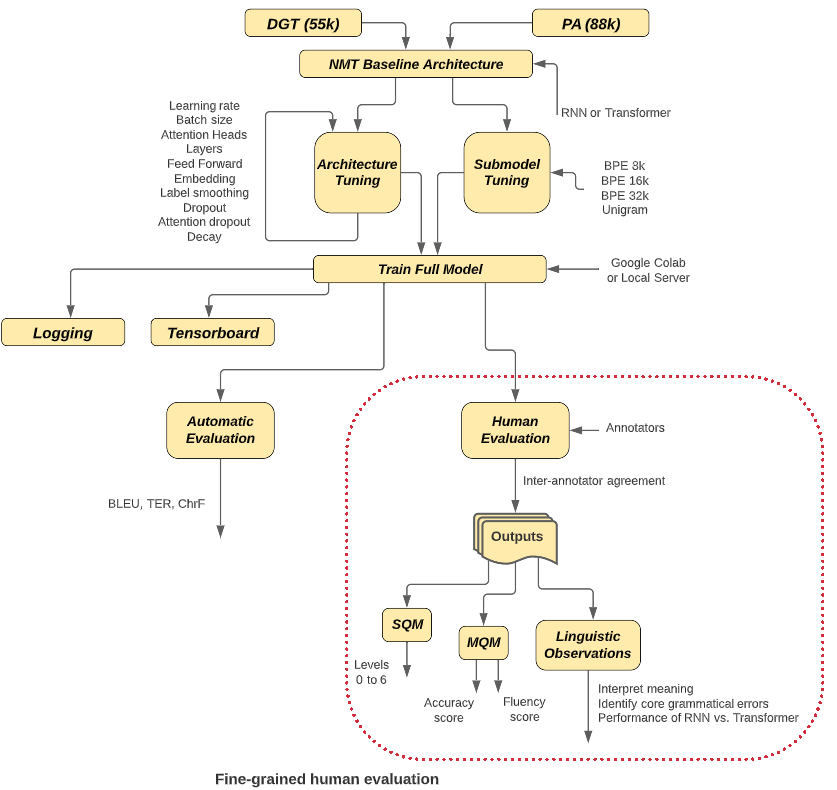}
\end{adjustwidth}
    \caption{The proposed approach to evaluate the baseline architectures of RNN and Transformer models is illustrated above. Using a random search approach, the values outlined in Table~\ref{tab:hpo-table} were tested to determine the optimal hyperparameters. Short cycles of 5k training steps were applied to test a range of values for each parameter. Once an optimal value was identified within the sampled range, it was locked in for tests on subsequent parameters. A fine-grained HE was conducted on the output from the DGT dataset and its results were compared with an automatic evaluation.}
    \label{fig:approach}
\end{figure}

In order to test the effectiveness of our approach, optimization was  carried out on an English--Irish parallel dataset: a general corpus of 52k lines from the Directorate General for Translation (DGT). With DGT, the test set used 1.3k lines and the development set comprised of 2.6k lines. All experiments involved concatenating source and target corpora to create a shared vocabulary and a shared SentencePiece subword model.  The adopted approach is illustrated in Figure \ref{fig:approach}.

\subsection{Architecture Tuning}

It is difficult and costly to tune systems using a conventional grid search approach given the long training times associated with NMT. Therefore, we adopted a random search approach in the HPO of our Transformer models. 

Using smaller and fewer layers with low-resource datasets has previously been shown to improve performance~\citep{araabi2020optimizing}. Furthermore, the use of shallow Transformer models has been demonstrated to improve the translation performance of low-resource NMT~\citep{van2020optimal}. Guided by these findings, configurations were tested, which varied the number of neurons in each layer and modified the number of layers used in the Transformer architecture.

Varying degrees of dropout were applied to Transformer models to evaluate the impact of regularization. Configurations using smaller (0.1) and larger values (0.3) were applied to the output of each feed-forward layer.

\subsection{Subword Models}

Incorporating a word segmentation approach, such as BPE, is now standard practice when developing NMT models. Subword models are particularly beneficial for low-resource languages since rare words are often a problem. In the context of English-to-Irish translation, there is no clear agreement as to what constitutes the best approach. Consequently, subword regularization techniques involving BPE and unigram models were evaluated as part of this study to determine the optimal parameters for maximizing translation performance. BPE models with varying vocabulary sizes of 4k, 8k, 16k and 32k were evaluated.

\begin{table}[H]
\small
\setlength{\tabcolsep}{18.4mm}
\caption{Transformer HPO using a random search approach. The optimal hyperparameters are highlighted in bold. The best-performing model used two attention heads and was trained on a 55k DGT corpus.}
\begin{tabular}{ll}
\toprule
\textbf{Hyperparameter} & \textbf{Values}                \\ \midrule
Learning rate            & 0.1, 0.01, 0.001, \textbf{{2}}            \\ 
Batch size               & 1024, \textbf{{2048}},  4096, 8192       \\
Attention heads          & \textbf{{2}}, 4, \textbf{{8}}              \\ 
Number of layers         & 5, \textbf{{6}}                           \\ 
Feed-forward dimension   & \textbf{{2048}}                           \\ 
Embedding dimension      & 128, \textbf{{256}}, 512                  \\ 
Label smoothing          & \textbf{{0.1}}, 0.3                       \\ 
Dropout                  & 0.1, \textbf{{0.3}}                       \\ 
Attention dropout        & \textbf{{0.1}}                            \\ 
Average Decay            & 0, \textbf{{0.0001}}                      \\ \bottomrule
\end{tabular}
\label{tab:hpo-table}
\end{table}

\subsection{Human Evaluation of NMT}

Morphological-rich languages, such as Irish, have a high degree of inflection and a free word order that gives rise to specific translation issues when translating from English. Grammatical categories such as gender or case inflections in nouns are often difficult to reliably generate in an Irish translation. 
\par
One of the goals of this research is to explore how an NMT system handles these issues compared with an RNN approach. Existing research suggests NMT systems should improve these linguistic aspects. NMT, with its use of subword models, implicitly addresses the problem in an unsupervised manner, without understanding the actual formal rules of grammatical categories.
\par
Previous HE studies that evaluate English--Irish MT performance have focused on the differences between an SMT and an NMT approach~\citep{dowling2018smt}. In the context of our research, HE was conducted on purely NMT methods, which included RNN and Transformer approaches. Furthermore, our study is differentiated by using both SQM and MQM as our HE metrics.

\par
It is clear from our earlier experimental findings, based solely on automatic evaluation metrics, that a Transformer approach leads to significant improvements compared to traditional RNN systems. However, as with most automatic scoring methods, these simply provide an overall score for each system but do not indicate the exact nature of the linguistic problems that may be encountered in translation. Therefore, it can be said that automatic evaluation does not address the question of the linguistic or grammatical quality of the target output. Nuances, such as how gender or cases are handled, are not covered by this~approach.    
\par
To achieve a deeper understanding of the linguistic errors created by our RNN and Transformer systems, a fine-grained HE was conducted. The outputs from these systems were systematically analyzed and compared in a manual error analysis. This approach captures the nature of the translation errors for each of the evaluated systems. The output from this study forms the basis of future work, which will help to improve the translation quality of our models. The annotation framework, the overall annotation process and inter-annotator agreement are discussed below, and broadly follow the approach adopted by other fine-grained HE studies \citep{klubivcka2018quantitative}.

\subsubsection{Scalar Quality Metrics}

SQM \citep{freitag2021experts} adapts the WMT shared-task settings to collect segment-level scalar ratings with a document context. SQM uses a scale from 0 to 6 for translation quality assessment. This is a modification of the WMT approach \citep{ma2017blend}, which uses a range from 0 to 100. 
 
 \clearpage
With this evaluation approach, annotators must select a rating from 0 through 6 when presented with the source and target sentences. The SQM quality levels for 0, 2, 4 and 6 are outlined in Table~\ref{tab:sqm}. Annotators may also choose intermediate levels of 1, 3 and 5 in cases where the translations do not exactly match the core SQM levels.

\begin{table}[H]
\small
\setlength{\tabcolsep}{8.16mm}
\caption{SQM levels explained \citep{freitag2021experts}.}

\begin{adjustwidth}{-\extralength}{0cm}
\centering 
\begin{tabular}{cp{13.5cm}}
\toprule
\multicolumn{1}{c}{\textbf{SQM Level}} &
  \textbf{Details of Quality} \\ \midrule
\multirow{2}{*}{ 6} &
  Perfect Meaning and Grammar: The meaning of the translation is completely consistent with the source and the surrounding context (if applicable). The grammar is also correct. \\ \midrule
\multirow{2}{*}{ 4} &
  Most Meaning Preserved and Few Grammar Mistakes: The translation retains most of the meaning of the source. This may contain some grammar mistakes or minor contextual inconsistencies. \\  \midrule
\multirow{2}{*}{ 2} & Some Meaning Preserved: The translation preserves some of the meaning of the source but misses significant parts. The narrative is hard to follow due to fundamental errors. Grammar may be~poor. \\  \midrule
 \multirow{2}{*}{ 0}&
  Nonsense/No meaning preserved: Nearly all information is lost between the translation and source. Grammar is irrelevant. \\ \bottomrule
\end{tabular}
\label{tab:sqm}
\end{adjustwidth}
\end{table}

\subsubsection{Multidimensional Quality Metrics}

As part of QTLaunchpad project (\url{https://www.qt21.eu/}, accessed June 2022).
the MQM framework (\url{https://www.qt21.eu/mqm-definition/definition-2015-12-30.html}, accessed June 2022) was developed to provide a framework of how manual evaluation could be performed via a detailed error analysis. A single metric for all uses is not imposed. Instead, a comprehensive catalogue of quality issue types, with standardized names and definitions, is provided. This catalogue may be customized for specific tasks. In addition to forming a reliable methodology for quality assessment, it also allows for us to specify which error tags were relevant to our task.
\par
\textls[-15]{To adapt the generic MQM framework for our context, we followed the official guidelines for scientific research \citep{lommel2018metrics}. The details of our customization of MQM are discussed~below}. 
\par
\begin{figure}[H]
    
\begin{adjustwidth}{-\extralength}{0cm}
 \centering
\includegraphics[width=17.8cm]{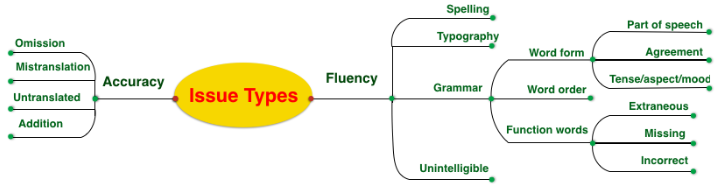}
\end{adjustwidth}
    \caption{The core set of error categories proposed by the MQM guidelines.}
    \label{fig:mqm_core}
\end{figure}

A large variety of tags, on several annotation layers, are proposed within the original MQM guidelines. However, this full MQM tagset is too detailed for a specific annotation task. Therefore, when evaluating our MT output, the smaller default set of evaluation categories, specified in the core tagset, were used. These standard top-level categories of accuracy and fluency, which are proposed by the MQM guidelines, are illustrated in Figure \ref{fig:mqm_core}. A special non-translation error was  used to tag an entire sentence, which was too badly translated to allow for the identification of individual errors.

Error severities are specified as either major or minor errors and are assigned independently of category. These correspond to actual translation/grammatical errors or smaller imperfections, respectively.  
The recommended default weights \citep{lommel2018metrics} were used, which allocate a weight of 1 to minor errors whereas major errors are assigned a weight of 10. Furthermore, the non-translation category was allocated a weight of 25, an approach which is line with the best practice established in previous studies \citep{freitag2021experts}. 

The annotators were instructed to identify all errors within each sentence of the translated output for both systems. The error categories used by the annotators are outlined in Table~\ref{tab:mqmcat}.

\begin{table}[H]
\small
\setlength{\tabcolsep}{4.6mm}
\caption{Description of error categories within the core MQM framework \citep{freitag2021experts}.}
			
\begin{adjustwidth}{-\extralength}{0cm}
\centering 
\begin{tabular}{p{2.7cm}p{2.7cm}p{10.25cm}}
\toprule
\textbf{Category} & \textbf{Sub-Category} & \textbf{Description}                                 \\ \midrule
\textbf{Non-translation} &                & Impossible to reliably characterize the 5 most severe errors.       \\ 
       \midrule
\textbf{Accuracy}        & Addition       & Translation includes information not present in the source.         \\
         & Omission              & Translation is missing content from the source.      \\
                & Mistranslation & Translation does not accurately represent the source.               \\
         & Untranslated text     & Source text has been left untranslated.              \\
        \midrule
\textbf{Fluency}  & Punctuation           & Incorrect punctuation                                \\
         & Spelling              & Incorrect spelling or capitalization.                \\
         & Grammar               & Problems with grammar, other than orthography.       \\
                         & Register       & Wrong grammatical register (e.g., inappropriately informal pronouns). \\
                  & Inconsistency         & Internal inconsistency (not related to terminology). \\
                  & Character encoding    & Characters are garbled due to incorrect encoding.   \\ \bottomrule
\end{tabular}

\label{tab:mqmcat}
\end{adjustwidth}
\end{table}

\subsubsection{Annotation Setup}

Annotations were carried using the simpler SQM approach and a more detailed, fine-grained MQM approach. The hierarchical taxonomy of our MQM implementation is illustrated in Figure \ref{fig:mqm_core}, whereas the SQM categories are summarized in Table~\ref{tab:sqm}. 
\par
Two annotators with similar backgrounds, were used for the annotation of  outputs from an RNN system and a Transformer system. Both annotators are native speakers of Irish and neither had prior experience with MQM. Prior to annotation, they were thoroughly familiarized with the process and the official MQM annotation guidelines. These guidelines offer detailed instructions for annotation within the MQM framework.
\par
Both annotators have been very involved in the education sector for decades. One of the annotators has edited numerous English-language and Irish-language books during her career as a university lecturer. The second annotator has a PhD in Irish-language place names. In addition, he has written numerous books in both English and Irish. Given their experience and strong language backgrounds, they were well-equipped to handle the task at hand.
\par
Using a test set of 20 randomly selected sentences, the annotators were presented with the English source text, an Irish reference translation and the two unannotated system outputs: one generated using an RNN model and the other created using a Transformer model. Potential bias was removed by using blind annotation such that annotators did not know which model the translation output came from. 
 The annotators worked independently of each other but were occasionally in contact to discuss the process and how to approach difficult sentences.
\par
Translations from the RNN and the Transformer system were annotated by both annotators, meaning that each system translated the same 20 sentences and each annotator annotated the resulting 40 translated sentences (20 source sentences for 2 MT systems), producing a total of 80 annotated sentences. The annotated dataset is publicly available on GitHub (\url{https://github.com/seamusl/isfeidirlinn}, accessed June 2022).

\par
Once the annotation data were extracted, each annotator analyzed the output to determine the performance of each system for each error category. 

\subsubsection{Inter-Annotator Agreement}

Low inter-annotator agreement (IAA) scores is a common problem experienced when using manual MT evaluation approaches such as MQM \citep{lommel2014using,  callison2007meta}.To determine the validity of the findings of our research, it is important to check the level of agreement between our annotators~\citep{artstein2017inter}. 

Cohen’s kappa ($k$) \citep{cohen1960coefficient} was used to determine inter-annotator agreement. Agreement was calculated based on the annotations of each individual system, with agreement being observed at the sentence level. With this approach, the differences in agreement across systems was explored and we also gained a high-level view of overall agreement between the annotators.  Furthermore, Cohen’s kappa was calculated separately for every error type and the findings are outlined in Table \ref{tab:my-cohen}.

\begin{table}[H]
\caption{Inter-annotator agreement using Cohen values}
\begin{tabular}{@{}clll@{}}
\toprule
\textbf{Error   type} & \multicolumn{1}{c}{\textbf{RNN}} & \multicolumn{1}{c}{\textbf{NMT}}  \\ \midrule
\multicolumn{1}{l}{Non-translation}               & 1.0 & 1.0    \\
\multicolumn{1}{l}{Accuracy}                      & 1.0 & 1.0   \\
Addition                                          & 1.0 & 1.0   \\
Omission                                          & 1.0 & 1.0   \\
Mistranslation                                    & -0.071 & 1.0   \\
Untranslated text                                 & 0.0 & 1.0   \\
\multicolumn{1}{l}{Fluency}                       &  &  &  \\
Punctuation                                       & 0.651 & 1.0    \\
Spelling                                          & 0.0 & 0.0    \\
Grammar                                           & 0.867 & 0.895   \\
Register                                          & 1.0 & 1.0   \\
Inconsistency                                     & 1.0 & 1.0   \\
Character Encoding                                & 1.0 & 1.0    \\ 
 \bottomrule
\end{tabular}

\label{tab:my-cohen}
\end{table}

\section{Empirical Evaluation\label{s4}}

\subsection{Experimental Setup}
\subsubsection{Datasets}
The performance of the Transformer and RNN approaches is evaluated on a publicly available English-to-Irish parallel dataset from the Directorate General for Translation (DGT)  (\url{https://ec.europa.eu/info/departments/translation}, accessed June 2022). The Joint Research Centre of the DGT has made all its translation memory (i.e. sentences and their professionally produced translations) available, which covers the official European Union languages~\citep{steinberger2013dgt}. Included in the training data are parallel texts from the Digital Corpus of the European Parliament (DCEP) and the DGT. Crawled data, from sites of a similar domain, are also incorporated. This dataset is broadly categorised as generic and is publicly~available.  

\subsubsection{Infrastructure}
Model development was conducted using local workstations, each of which was built with an AMD Ryzen 7 2700X processor, 16 GB memory, a 256 SSD and an NVIDIA GeForce GTX 1080 Ti. 

In addition, a Google Colab Pro subscription enabled rapid prototype development and created zero-emission models. The available computing power of the Google Cloud was much higher than our local infrastructure and provided servers with 16 GB graphic cards (NVIDIA Tesla P100 PCIe) and up to 27~GB of memory~\citep{bisong2019google}. Larger Transformer models were built on local infrastructure, since long builds timed out on Colab due to Google restrictions. The Pytorch implementation of OpenNMT 2.0, an open-source toolkit for NMT~\citep{klein2017opennmt}, was used to train all MT models.

\subsubsection{Metrics}

The performance of all models was evaluated using the automated metrics of BLEU~\citep{papineni2002BLEU}, TER~\citep{snover2006study} and ChrF~\citep{popovic2015ChrF}. Case-insensitive BLEU scores are reported at the corpus level.  

\subsection{Automatic Evaluation Results}

\subsubsection{Performance of Subword Models}
The impact that choice of subword model has on translation is highlighted in Tables \ref{tab:dgtvanilla-table} and  \ref{tab:trans-table}. Incorporating any subword model type led to improvements in model accuracy when training both RNN and Transformer architectures.  

\begin{table}[H]
\small
\setlength{\tabcolsep}{6.65mm}
\caption{RNN performance on DGT dataset of 52k lines. There were zero carbon emissions in building these models, since smaller RNN models were trained on Google Colab servers, which are carbon-neutral.}

\begin{adjustwidth}{-\extralength}{0cm}
\centering 
\begin{tabular}{lcccccccc}
\toprule
\textbf{Architecture} &
  \textbf{BLEU} \boldmath{$\uparrow$} &
  \textbf{TER} \boldmath{$\downarrow$} &
  \textbf{ChrF3} \boldmath{$\uparrow$} &
  \textbf{Steps} &
  \textbf{Runtime  (h)} &
  \textbf{kgCO\textsubscript2} \\ \midrule
dgt-rnn-base    & 52.7       & 0.42  & 0.71 & 75k  & 4.47 & 0 \\
dgt-rnn-bpe8k   & 54.6       & 0.40 & 0.73 & 85k  & 5.07 & 0 \\
dgt-rnn-bpe16k  & 55.6 & 0.39 & 0.74 & 100k & 5.58 & 0 \\
dgt-rnn-bpe32k  & 55.3       & 0.39 & 0.74 & 95k  & 4.67 & 0 \\ 
dgt-rnn-unigram & 55.6 & 0.39 & 0.74 & 105k & 5.07 & 0 \\ \bottomrule 
\end{tabular}

\label{tab:dgtvanilla-table}
\end{adjustwidth}
\end{table}

\vspace{-9pt}

\begin{table}[H]
\small
\setlength{\tabcolsep}{6.5mm}
\caption{Transformer performance on 52k DGT dataset. The highest performing model uses 2 attention heads. All other models use 8 attention heads. Transformer models were long-running builds, which had to be carried out on local servers.}

\begin{adjustwidth}{-\extralength}{0cm}
\centering 

\begin{tabular}{lcccccccc}
\toprule
\textbf{Architecture} &
  \textbf{BLEU} \boldmath{$\uparrow$} &
  \textbf{TER} \boldmath{$\downarrow$} &
  \textbf{ChrF3} \boldmath{$\uparrow$} &
  \textbf{Steps} &
  \textbf{Runtime (h)} &
  \textbf{kgCO\textsubscript2} \\ \midrule
dgt-trans-base      & 53.4 & 0.41 & 0.72 & 55k  & 14.43 & 0.81 \\
dgt-trans-bpe8k     & 59.5 & 0.34 & 0.77 & 200k & 24.48 & 1.38 \\
dgt-trans-bpe16k    & 60.5 & 0.33 & 0.78 & 180k & 26.90 & 1.52 \\
dgt-trans-bpe32k    & 59.3 & 0.35 & 0.77 & 100k & 18.03 & 1.02 \\ 
dgt-trans-unigram   & 59.3 & 0.35  & 0.77 & 125k & 21.95 & 1.24 \\ \bottomrule
\end{tabular}

\label{tab:trans-table}
\end{adjustwidth}
\end{table}

A baseline RNN model, illustrated in Table \ref{tab:dgtvanilla-table}, achieved a BLEU score of 52.7, whereas the highest-performing BPE variant,  using a 16k vocab, recorded an improvement of nearly three points, with a score of 55.6. 

In the context of Transformer architectures, highlighted in Table \ref{tab:trans-table}, the use of subword models delivers significant performance improvements. The performance gains for Transformer models are much higher compared with the improvements recorded by the RNN models. A baseline Transformer model achieves a BLEU score of 53.4, whereas a Transformer model, with a 16k BPE submodel, has a score of 60.5, representing a BLEU score improvement of 13\% at 7.1 BLEU points. 

For translating into a morphologically rich language, such as Irish, the ChrF metric has proven successful in showing a strong correlation with human translation~\citep{stanojevic2015results}. In the context of our experiments, this worked well in highlighting the performance differences between RNN and Transformer architectures. 

\subsubsection{Transformer Performance Compared with RNN}

\begin{figure}[H]
    \includegraphics[width=13.5cm]{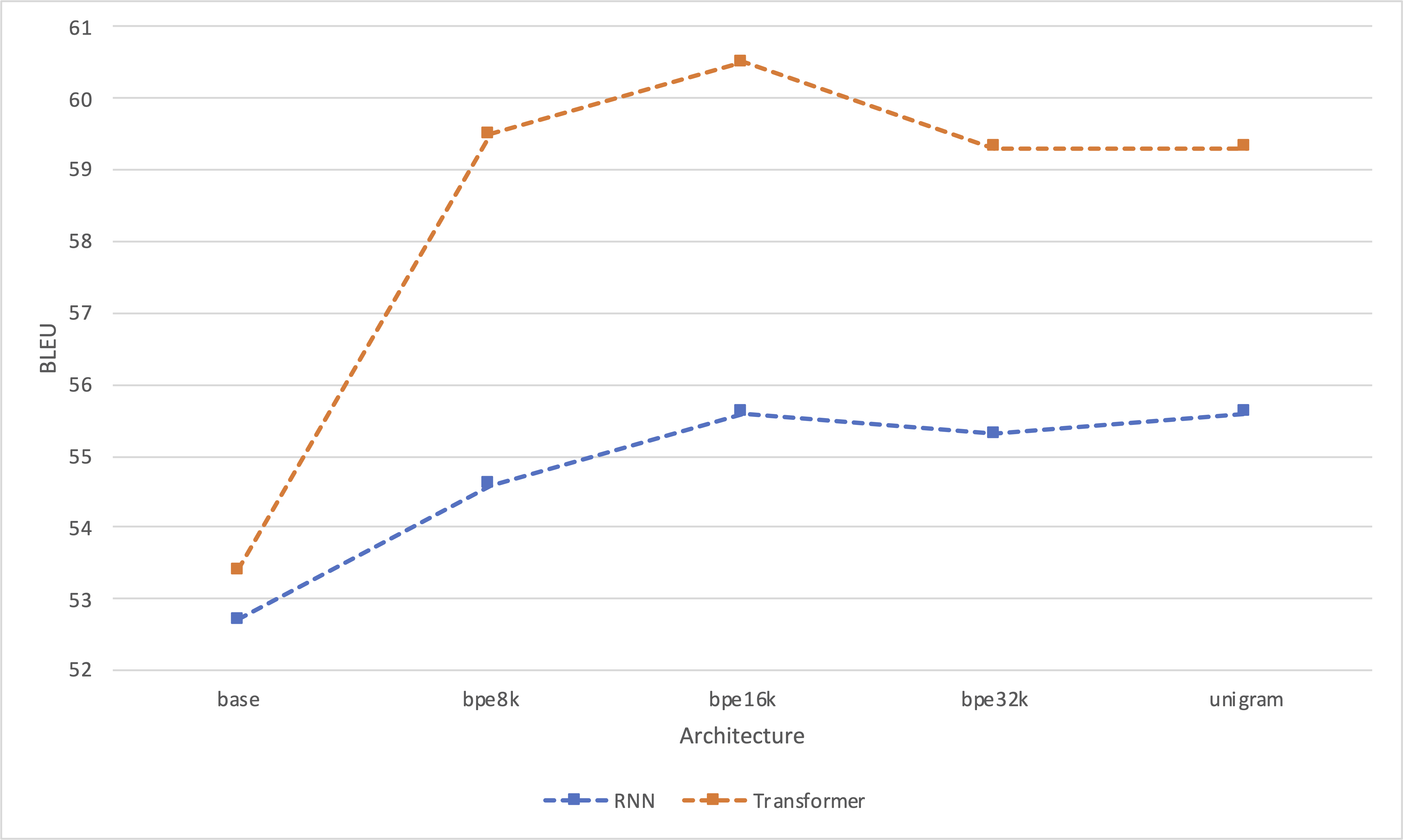}
    \caption{BLEU performance for all model architectures is compared. The use of a BPE subword model improved translation performance in all cases. The best-performing model was built using a 16k BPE subword model on a Transformer architecture. }
    \label{fig:dgt}
\end{figure}

\begin{figure}[H]
    \includegraphics[width=13.5cm]{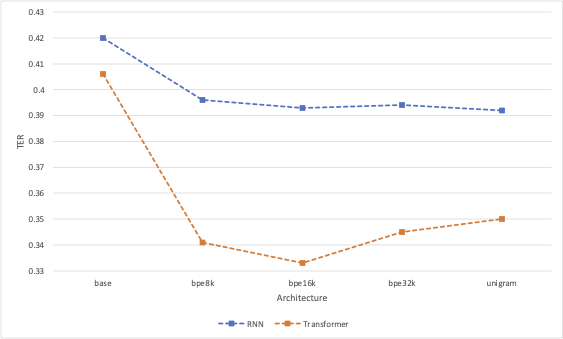}
    \caption{TER performance for all model architectures. The highest-performing model uses a 16k BPE subword model on a Transformer architecture. In all instances, incorporating a subword model improves TER.}
    \label{fig:pacompare}
\end{figure}

The performance of RNN models is contrasted with the Transformer approach in \linebreak  Figures \ref{fig:dgt} and \ref{fig:pacompare}. Transformer models, as anticipated, outperformed all their RNN counterparts. It is interesting to note the impact of choosing the optimal vocabulary size for BPE submodels. Choosing a BPE vocabulary of 16k words yields the highest performance. 

Furthermore, the TER scores highlighted in Figure \ref{fig:pacompare} reinforce the findings that using 16k BPE submodels on Transformer architectures leads to a better translation performance. The TER score for the 16k BPE Transformer model is significantly better (0.33) when compared with the baseline performance (0.41).

\vspace{-9pt}

\subsection{Human Evaluation Results}

The aggregate total of errors found by annotators for each system is highlighted in Table~\ref{tab:mqmtotals}. Looking at the aggregate data alone, it is evident that both annotators have judged that the RNN system contains more errors, and that the NMT system contains less errors.

\begin{table}[H]
\small
\setlength{\tabcolsep}{9.06mm}
\caption{Total errors found by each annotator using the MQM metric.}
\begin{tabular}{@{}lcccc@{}}
\toprule
                      & \multicolumn{2}{c}{\textbf{Annotator 1}} & \multicolumn{2}{c}{\textbf{Annotator 2}} \\ \midrule
\textbf{System}       & RNN                 & Transformer                & RNN                 & Transformer                \\ \midrule
\textbf{Total Errors} & 41                     & 23                    & 43                     & 23                   \\ \bottomrule
\end{tabular}

\label{tab:mqmtotals}
\end{table}

\par
While such a high-level view is instructive in determining which system is better, it lacks the granularity required to pinpoint the linguistic aspects of how these translations can be improved. To achieve a deeper insight, a fine-grained analysis of the error types was conducted, the results of which are displayed in Table~\ref{tab:combined}. Categorized by error type, the sum of error tags by each annotator for each system is outlined.

\begin{table}[H]
\small
\setlength{\tabcolsep}{17.72mm}
\caption{Transformer and RNN approach is compared using concatenated annotation data across both annotators. In all MQM error categories, the Transformer architecture performs better, apart from a tie in the omission category.}
\begin{tabular}{@{}lcccc@{}}
\toprule
\multicolumn{1}{l}{} &
  \multicolumn{1}{c}{\textbf{RNN}} &
  \multicolumn{1}{c}{\textbf{NMT}} \\ \midrule
\textbf{Error Type} &
  \multicolumn{1}{c}{\textbf{Error}} &
  \multicolumn{1}{c}{\textbf{Error}} \\ \midrule
Non-translation       & 0 & 0    \\
\textbf{Accuracy}              &  &     \\
Addition                                  & 10 & 4   \\
Omission                                  & 12 & 12   \\
Mistranslation                            & 26  & 14    \\
Untranslated text                         & 4 & 1   \\
\textbf{Fluency}               &  &    \\
Punctuation                               & 5 & 4   \\
Spelling                                  & 1 & 0   \\
Grammar                                   & 20 & 11   \\
Register                                  & 2 &  0  \\
Inconsistency                             & 2 & 0   \\
Character Encoding                        & 0 & 0   \\ \midrule
\multicolumn{1}{l}{\textbf{Total errors}} & 82 & 46    \\ \bottomrule
\end{tabular}

\label{tab:combined}
\end{table}

\section{Environmental Impact\label{s5}}
The environmental impact of all aspects of computing has received increased research interest in recent times. Much of this effort has concentrated on NMT's carbon footprint~\citep{jooste2022knowledge, bender2021dangers}. To assess the environmental impact of our NMT models, we tracked energy consumption during their development. 

Prototype model development was carried out using Google Colab which is a carbon neutral platform~\citep{lacoste2019quantifying}. However, longer running Transformer experiments were conducted on local servers using 324 gCO\textsubscript2 per kWh (\url{https://www.seai.ie/publications/Energy-in-Ireland-2020.pdf}, accessed June 2022) \citep{sei2020}. The net result was just under 10 kgCO\textsubscript2, created for a full run of model development. Models developed during this study will be reused for ensemble experiments in future work.

The environmental costs of our model development were tracked to serve as a benchmark for future work. Awareness of such costs will impose a discipline on our work, such that we opt for carbon-neutral cloud providers. In cases where models are developed on local infrastructure, this will encourage the use of more efficient GPUs and the utilization of techniques that result in faster builds.

\section{Discussion\label{s6}}

\begin{figure}[H]

    \includegraphics[width=7cm]{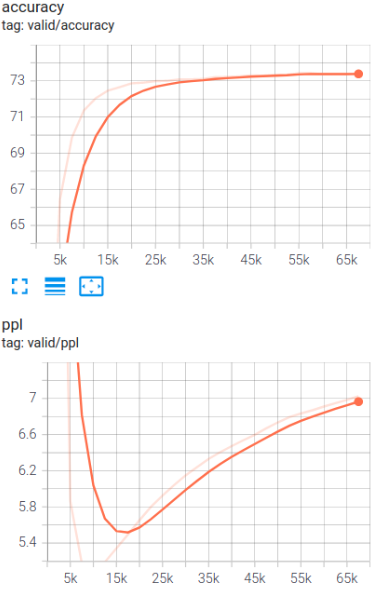}
    \caption{Transformer baseline.}
    \label{fig:train-base}
\end{figure}

\begin{figure}[H] 

    \includegraphics[width=7cm]{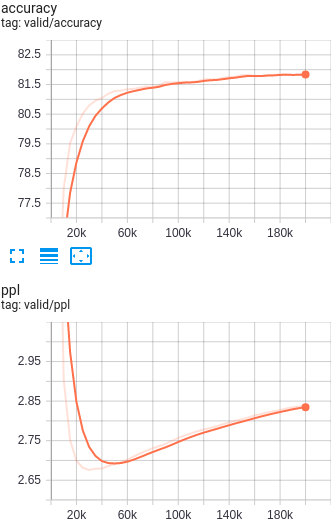}
    \caption{Transformer 16k BPE subword model.}
    \label{fig:train-bpe}
\end{figure}

Validation accuracy and model perplexity (PPL) in developing the baseline and optimal Transformer models are illustrated in Figures \ref{fig:train-base} and  \ref{fig:train-bpe}. Training a Transformer model with a 16k BPE subword model boosted the validation accuracy by over 8\% compared to its baseline.

\vspace{-10pt}

Rapid convergence was observed while training the baseline model such that little accuracy improvement occurs after 20k steps. Including a subword model led to slower converging models, with only marginal gains recorded after 60k steps. Examining \linebreak Figures \ref{fig:train-base} and  \ref{fig:train-bpe}, PPL achieves a lower global minimum when the Transformer approach is used with a 16k BPE submodel. The PPL global minimum (2.7) is over 50\% lower than the corresponding PPL for the Transformer base model (5.5). This finding illustrates that choosing an optimal subword model delivers significant performance gains.

Translation engine performance, at the corpus level, was benchmarked against Google Translate's (\url{https://translate.google.com/}, accessed June 2022) English-to-Irish translation service, which is freely available on the internet. Four random samples were selected from the English source test file and are presented in Table~\ref{tab:translations}. Translation of these samples was carried out on the optimal Transformer model using Google Translate. Case-insensitive, sentence-level BLEU scores were recorded and are presented in Table~\ref{tab:trans-google}. It must be acknowledged that this comparison is not entirely valid given that Google does not have access to our training data, nor do we have unlimited access to the Google cloud infrastructure. Nonetheless, the results are encouraging and indicate a good performance by our translation models on the DGT dataset.
 
 \begin{table} [H]
\small
\setlength{\tabcolsep}{13.88mm}
\caption{Random samples of human reference translations taken from the test dataset.}
\begin{small}

\begin{adjustwidth}{-\extralength}{0cm}
\centering 
\begin{tabular}{ll}

\toprule
\textbf{Source Language (English)} & \textbf{Reference Human Translation (Irish)} \\  \midrule
\begin{tabular}[c]{@{}l@{}}A clear harmonised procedure, including the \\ necessary criteria for disease–free status, \\ should be established for that purpose.\end{tabular} & \begin{tabular}[c]{@{}l@{}}Ba cheart nós imeachta comhchuibhithe soiléir, \\ lena n-áirítear na critéir is gá do stádas saor \\ ó ghalar, a bhunú chun na críche sin.\end{tabular} \\ \midrule
\begin{tabular}[c]{@{}l@{}}the mark is applied anew, as appropriate.\end{tabular} & \begin{tabular}[c]{@{}l@{}}déanfar an mharcáil arís, mar is iomchuí.\end{tabular} \\ \midrule
\begin{tabular}[c]{@{}l@{}}If the court decides that a review is \\ justified on any of the grounds set out in \\ paragraph 1, the judgment given in the \\ European Small Claims Procedure shall \\ be null and void.\end{tabular} & \begin{tabular}[c]{@{}l@{}}Má chinneann an chúirt go bhfuil bonn cirt \\ le hathbhreithniú de bharr aon cheann de na \\ forais a leagtar amach i mír 1, beidh an \\ breithiúnas a tugadh sa Nós Imeachta Eorpach \\ um Éilimh Bheaga ar neamhní go hiomlán.\end{tabular} \\ \midrule
households where pet animals are kept; & teaghlaigh ina gcoimeádtar peataí; \\ \bottomrule

\end{tabular}
\end{adjustwidth}

\label{tab:translations}
\end{small}

\end{table}
\vspace{-9pt}

\begin{table}[H]
\small
\setlength{\tabcolsep}{8.2mm}
\caption{Transformer model compared with Google Translate using random samples from the DGT corpus. Full evaluation of Google Translate's engines on the DGT test set, with 1.3k lines, generated a BLEU score of 46.3 and a TER score of 0.44. Comparative scores on the test set using our Transformer model, with 2 attention heads and 16k BPE submodel realised 60.5 for BLEU and 0.33 for TER.}
\begin{small}

\begin{adjustwidth}{-\extralength}{0cm}
\centering 
\begin{tabular}{lclc}
\toprule
\textbf{Transformer (16k BPE)} & \textbf{BLEU} \boldmath{$\uparrow$} & \textbf{Google Translate} & \textbf{BLEU} \boldmath{$\uparrow$}\\ \midrule

\begin{tabular}[c]{@{}l@{}}Ba cheart nós imeachta soiléir \\ comhchuibhithe, lena n-áirítear \\ na critéir is gá maidir le \\ stádas saor ó ghalair, a bhunú \\ chun na críche sin.
\end{tabular} & 61.6 & \begin{tabular}[c]{@{}l@{}}Ba cheart nós imeachta \\ comhchuibhithe soiléir, lena \\ n-áirítear na critéir riachtanacha \\ maidir le stádas saor ó ghalair, \\ a bhunú chun na críche sin.\end{tabular} & 70.2 \\\midrule
\begin{tabular}[c]{@{}l@{}}go gcuirtear an marc i bhfeidhme,\\ de réir mar is iomchuí.\end{tabular} & 21.4 & \begin{tabular}[c]{@{}l@{}}cuirtear an marc i bhfeidhm as \\ an nua, de réir mar is cuí.\end{tabular} & 6.6 \\ \midrule
\begin{tabular}[c]{@{}l@{}}
Má chinneann an chúirt go bhfuil \\ bonn cirt le hathbhreithniú ar aon \\ cheann de na forais a leagtar amach \\ i mír 1, beidh an breithiúnas a \\ thugtar sa Nós Imeachta Eorpach \\ um Éilimh Bheaga ar neamhní.

\end{tabular} & 77.3 & \begin{tabular}[c]{@{}l@{}}Má chinneann an chúirt go bhfuil \\ údar le hathbhreithniú ar aon \\ cheann de na forais atá leagtha \\ amach i mír 1, beidh an \\ breithiúnas a thugtar  sa \\ Nós Imeachta Eorpach um \\ Éilimh Bheaga ar neamhní\end{tabular} & 59.1 \\ \midrule
teaghlaigh ina gcoimeádtar peataí; & 100 & teaghlaigh ina gcoinnítear peataí; & 30.2 \\ \bottomrule
\end{tabular}

\label{tab:trans-google}
\end{adjustwidth}
\end{small}

\end{table}

The optimal parameters selected in this discovery process are identified in bold in Table~\ref{tab:hpo-table}. A higher initial learning rate of 2 coupled with an average decay of 0.0001 led to longer training times but more accurate models. Despite setting an early stopping parameter, many of the Transformer builds continued for the full cycle of 200k steps over periods of 20+ hours. 

Training Transformer models with a reduced number of attention heads led to a marginal improvement in translation accuracy with a smaller corpus. Our best-performing model achieved a BLEU score of 60.5 and a TER score of 0.33 with 2 heads and a 16k BPE submodel. By comparison, using 8 heads with the same architecture and dataset yielded 60.3 for BLEU and 0.34 in terms of TER. 

Transformer models developed, using state-of-the-art techniques, were evaluated as part of the LoResMT2021 Shared Task \citep{ojha2021findings}. Models developed using our approach, as outlined above, were entered into the competition, and the highest-performing EN-GA system was submitted by our team (ADAPT) \citep{lankford2021machine}.

\subsection{Inter-Annotator Reliability}

In Cohen's original article \citep{cohen1960coefficient}, the interpretation of specific $k$ scores is clearly outlined. There is no agreement with values $\le$ 0, none to slight agreement when scores are in the range of 0.01--0.20, fair agreement is represented by 0.21--0.40, 0.41--0.60 is moderate agreement, 0.61--0.80 is substantial agreement, and 0.81--1.00 is almost perfect agreement. 

The literature \citep{mchugh2012interrater} recommends a minimum of 80\% agreement for good inter-annotator agreement. As illustrated in Table \ref{tab:my-cohen}, there is almost perfect agreement between the annotators when evaluating output from the NMT models. In the case of the RNN outputs, there is disagreement in the mistranslation category but agreement in all other categories. Given these scores, we have a high degree of confidence in our human evaluation of both the RNN and NMT outputs.

\subsection{Performance of Is Féidir Linn Models Relative to Google}

Using standard Transformer parameters, such as a  batch size of 2048 and setting the number of encoder/decoder layers to 6, were observed to perform well. Increasing the regularization dropout to 0.3 and reducing hidden neurons to 256 improved translation performance. Consequently, these values were selecting when building all Transformer~models.

\subsection{Linguistic Observations}

A linguistic analysis of the outputs from the Transformer-optimized model is illustrated in Table \ref{tab:ling-anal}. The English language source sentences and their Irish language translations are presented. The sentences have been selected from fine-grained human evaluation, since they highlight some of the key error types that are encountered. The analysis focuses on the shortcomings of our model outputs, which fall into the following categories: interpretative meaning, core grammatical errors and commonly used irregular verbs. Finally, using the HE metrics of SQM and MQM, the performance of an RNN approach is contrasted with that of the Transformer approach.  

\subsubsection{Interpreting Meaning}

The generic Irish verb ``déan'' (to do or to make) is used to express more precise concepts such as ``to conduct'', ``to put into effect'' or ``to carry out''. Both the RNN and Transformer systems make use of ``déan'' in a generic way, but they fail to capture the refinement of concept expressed in each of these meanings. An example of this problem is illustrated in GA-1 in Table~\ref{tab:ling-anal}. In this context, a more natural and intuitive translation to capture the expression ``to conduct'' would be to substitute ``a dhéanamh'' with \linebreak ``a sheoladh''.

\clearpage
A similar lack of refinement from both systems is also found with the usage of other words. For example, ``cuid'' (part) is used to translate ``operative part'' in GA-2. However, a more precise interpretation would be the usage of ``gné'', leading to the correct translation ``gné oibríochtúil'' i.e., ``operative part''.

Another example where the translation models failed to correctly interpret the true sense of an English source word into a corresponding Irish translation can be seen in GA-3. The Irish verb ``Mainnigh'' meaning ``to default'' would not be used in the context of the source text in EN-3. Using the Irish verb ``teip'', meaning ``to fail'', is the correct translation of the idea ``fails to meet the performance requirements'': ``má theipeann an t-oibreoir na ceanglais feidhmíochta a chomhlíonadh.''  This error was observed in both the RNN and Transformer model outputs.

\begin{table}[H]
\small
\setlength{\tabcolsep}{4.85mm}
\caption{Linguistic analysis of system outputs. Sources of errors are flagged in blue and in red.}

\begin{adjustwidth}{-\extralength}{0cm}
\centering 
\begin{tabular}{p{2cm}p{14.5cm}}
\toprule
\textbf{Type} &
  \textbf{Sentence} \\ \midrule 
\textbf{EN-1} & The lead supervisory authority   may request at any time other supervisory authorities concerned to provide   mutual assistance pursuant to Article 61 and \textcolor{blue}{may conduct} joint operations pursuant to Article 62, in particular for carrying out investigations or for monitoring the implementation of a measure concerning a controller or   processor established in another Member State. \\ \midrule 
\textbf{GA-1} &
  Féadfaidh an príomhúdarás  maoirseachta iarraidh, tráth ar bith, ar bith eile lena mbaineann cúnamh   frithpháirteach a chur ar fáil de bhun Airteagal 61 agus féadfaidh sé oibríochtaí comhpháirteacha a dhéanamh de bhun Airteagal 62, go háirithe   maidir le himscrúduithe a dhéanamh nó maidir le faireachán \textcolor{red}{a dh\'{e}anamh} ar chur chun feidhme beart i ndáil le rialaitheoir nó próiseálaí atá bunaithe i mBallstát eile. \\\midrule
\textbf{EN-2} &
  The Office shall mention the judgment in the Register and shall take the necessary measures to comply with its operative \textcolor{blue}{part}. \\ \midrule
\textbf{GA-2} &
 Luafaidh an Oifig an breithiúnas   sa Chlár agus glacfaidh sí na bearta is gá chun cloí lena \textcolor{red}{chuid} oibríochtúil. \\ \midrule
\textbf{EN-3} &
The competent authority may at any time wholly or partially suspend or terminate the contract awarded under this provision if the operator \textcolor{blue}{fails} to meet the performance requirements. \\\midrule
\textbf{GA-3} &
Féadfaidh an t-údarás inniúil an conradh a dámhadh faoin bhforáil seo a chur ar fionraí nó a fhoirceannadh go hiomlán nó go páirteach \textcolor{red}{m\'{a} mhainn\'{i}onn} an t-oibreoir na ceanglais feidhmíochta a chomhlíonadh. \\ \midrule
\textbf{EN-4} &
This Directive shall enter into force on the day following that of its \textcolor{blue}{publication} in the Official Journal of the European Union. \\ \midrule
\textbf{GA-4} &
Tiocfaidh an Treoir seo i bhfeidhm an lá tar éis lá \textcolor{red}{a fhoilsithe} in Iris Oifigiúil an Aontais Eorpaigh. \\ \midrule
\textbf{EN-5} &
Such special measures are interim in nature, and \textcolor{blue}{shall not be} subject to the conditions set out in Article 7(1) and (2). \\ \midrule
\textbf{GA-5} &
Tá bearta speisialta den sórt sin eatramhach, agus \textcolor{red}{ní bheidh said} faoi réir na gcoinníollacha a leagtar amach in Airteagal 7(1) agus (2) iad. \\\bottomrule
\end{tabular}
\end{adjustwidth}

\label{tab:ling-anal}
\end{table}

\subsubsection{Core Grammatical Errors}

Grammatical mistakes in the form of the misuse of lenitions (e.g., GA-4), incorrect pronouns (e.g., GA-5) and register errors (e.g., GA-5) were observed in both translation architectures. However, as is evident from both the automatic and MQM evaluations, there were far fewer errors with the Transformer model. Evidence of this can be seen in Table~\ref{tab:ling-anal}. In the case of GA-4, the RNN model included the lenition in ``a foilsithe'', whereas the Transformer model correctly removed ``h''. The correct use of the feminine noun ``treoir'' requires the removal of ``h'' in ``fhoilsithe''. 

The misuse of pronouns was observed in the RNN translation model and, to a lesser degree, in the Transformer model. In the case of GA-5, the RNN's incorrect use of the pronoun ``ní bheidh siad'' (they will not) is illustrated, whereas the Transformer approach used the correct form ``ní bheidh sé'' (he will not).  

\clearpage

Within the same sentence, GA-5, there is also evidence of a register error. In the English source text EN-5, the use of ``shall not be subject to'' expresses a stipulation. This is not registered in the Irish translation of ``ní bheidh said'', which simply, and less forcefully, means ``they will not''. This incorrect use of register was observed with both the RNN and the Transformer approaches. A more formal and closer interpretation of the English source would be the use of the imperative mode: ``ná bídís'' (let it not be).

\subsubsection{Commonly-Used Irregular Verbs}

One of the main inadequacies observed in both the RNN and Transformer systems is a lack of refinement of verbal usage, particularly when using the verbs ``déan'' (to do or to make ) and ``bí'' (to be). As in many languages, the fact that these are possibly the two most universally used verbs in Irish further exacerbates the problem. An illustration of this problem can be seen in the output GA-1, which highlights the incorrect usage of  ``déan''. In a similar fashion, GA-5 demonstrates how the system misinterprets the usage of the verb ``bí'', e.g.,  ``ní bheidh said''.

\subsubsection{Performance of RNN Approach Relative to Transformer Approach}

There is a strong correlation between automatic and human evaluation of the translation systems that we developed. The automatic BLEU scores are contrasted with the HE scores for both the RNN and Transformer models in Table~\ref{tab:all-metrics}. 

\begin{table}[H]
\small
\setlength{\tabcolsep}{10.85mm}

\caption{Transformer approach compared to the RNN approach across all metrics for the DGT dataset. The results from our HE, using SQM and MQM metrics, validate the BLEU automatic evaluation results.}
\begin{tabular}{@{}llccc@{}}
\toprule
\textbf{Approach}    &  & \textbf{BLEU} \boldmath{$\uparrow$} & \textbf{SQM} \boldmath{$\uparrow$} & \textbf{MQM} \boldmath{$\uparrow$} \\ \midrule
\textbf{Transformer} &  & 60.5          & 4.53         & 77.92        \\
\textbf{RNN}         &  & 52.7          & 3.30         & 43.05        \\ \bottomrule
\end{tabular}

\label{tab:all-metrics}
\end{table}

\subsection{Limitations of the Study}
Certain aspects of this study could be further developed, given more time and resources. Although there is high inter-annotator agreement, it would help to have more annotators. In addition, the human evaluation of a greater number of lines, coupled with a more detailed MQM taxonomy, may provide greater insight into the MT outputs. This would help in uncovering other aspects, such as how gender is handled by the MT models.

\section{Conclusions and Future Work\label{s7}}

With this research, we have presented the first HE study that compares the output of EN-GA RNN systems with that of Transformer-based EN-GA systems. Automatic metrics were shown to differentiate the systems and highlighted that Transformer models are superior to RNN models. In our paper, we demonstrated that a random search approach to HPO enabled the development of high-performing translation models. We have shown there is a high level of correlation between an HE and an automatic approach. Both the automatic metrics and our HE demonstrated that the Transformer-based system is the most~accurate. 

The importance of selecting hyperparameters when training low-resource Transformer models was also demonstrated. By increasing dropout and reducing the number of hidden-layer neurons, our models performed significantly better than Google Translate and our baseline models.

We have demonstrated that choosing the correct subword models is an important performance driver for low-resource MT. Within the context of low-resource English-to-Irish translations, we achieved optimal performance on a 55k generic corpus when a Transformer architecture with a 16k BPE subword model was used. Improvements in the performance of our optimized Transformer models was observed across all key indicators, namely, PPL was achieved at a lower global minimum, with a lower post-editing effort and a higher translation~accuracy.

As part of future work, steps can be taken to deal with the inadequacies highlighted in our linguistic analysis. The issue of misusing common irregular verbs could be addressed by fine-tuning our models with a dataset specifically tailored for that purpose.  In a similar fashion, fine-tuning after the careful selection of training data would also reduce the register errors encountered in our linguistic analysis. As it is difficult to train systems for all eventualities, using post-editing tools would be the best approach to correcting core grammatical errors involving pronouns, lenitions and lemmatization.

\vspace{6pt}

\authorcontributions{All authors have read and agreed to the published version of the manuscript.}

\funding{This work was supported by ADAPT, which is funded under the SFI Research Centres Programme (Grant 13/RC/2016) and is co-funded by the European Regional Development Fund. This research was also funded by the Munster Technological University.}

\informedconsent{Informed consent was obtained from all subjects involved in the study.}

\dataavailability{The data presented in this study are openly available at \url{https://github.com/seamusl/isfeidirlinn} }

\acknowledgments{We would like to thank the annotators, Dr Éamon Lankford and Ms. Máirín Lankford for their meticulous work in annotating the system outputs. }

\conflictsofinterest{ The authors declare no conflict of interest. The funders had no role in the design of the study; in the collection, analyses, or interpretation of data; in the writing of the manuscript, or in the decision to publish the~results.} 

\abbreviations{Glossary}{Irish terms referenced and used in this manuscript:\\
\noindent 
\begin{tabular}{@{}ll}
Déan & To do or to make\\
Bí & To be \\ 
Ná bídís & Let it not be\\
Ní bheidh siad & They will not\\
Ní bheidh sé & He will not \\
\end{tabular}}


\begin{adjustwidth}{-\extralength}{0cm}

\reftitle{References}

\end{adjustwidth}
\end{document}